\documentclass[10pt,twocolumn,letterpaper]{article}

\usepackage{times}
\usepackage{epsfig}
\usepackage{graphicx}
\usepackage{amsmath}
\usepackage{amssymb}
\usepackage{makecell, multirow}

\usepackage[pagebackref=true,breaklinks=true,letterpaper=true,colorlinks,bookmarks=false]{hyperref}

\begin{document}

\title{Reconstructing Seen Image from Brain Activity by Visually-guided Cognitive Representation and Adversarial Learning }

\author{Ziqi Ren\\
Xidian University\\
\and
 Jie Li\\
 Xidian University\\
\and
Xuetong Xue\\
Xidian University\\
\and
Xin Li\\
IIAI\\
\and
Fan Yang\\
IIAI\\
\and
Zhicheng Jiao\\
UPenn\\
\and
Xinbo Gao\\
Xidian University\\
}

\maketitle

\begin{abstract}
Reconstructing visual stimulus (image) \emph{only} from human brain activity measured with functional Magnetic Resonance Imaging (fMRI) is a significant and meaningful task in Human-AI collaboration. However, the inconsistent distribution and representation between fMRI signals and visual images cause the heterogeneity gap. Moreover, the fMRI data is often extremely high-dimensional and contains a lot of visually-irrelevant information. Existing methods generally suffer from these issues so that a satisfactory reconstruction is still challenging. In this paper, we show that it is possible to overcome these challenges by learning visually-guided cognitive latent representations from the fMRI signals, and inversely decoding them to the image stimuli. The resulting framework is called Dual-Variational Autoencoder/ Generative Adversarial Network ({\scshape{D-Vae/Gan}}), which combines the advantages of adversarial representation learning with knowledge distillation. In addition, we introduce a novel three-stage learning approach which enables the (cognitive) encoder to gradually distill useful knowledge from the paired (visual) encoder during the learning process. Extensive experimental results on both artificial and natural images have demonstrated that our method could achieve surprisingly good results and outperform all other alternatives.

\end{abstract}

\section{Introduction}
Reading mind, the act of ``seeing'' someone’s thoughts and feelings, has long been an ambitious capability in works of fiction. In recent years, breakthroughs in neuroscience and AI have brought such fictional technologies into the realm of science, \emph{i.e.}, brain decoding. Neuroscience studies~\cite{a48} have suggested that there exist a mapping from visual stimuli to brain activity patterns, which takes a visual stimulus as input and produces the corresponding brain activity pattern. Recent studies~\cite{a49} find that such mapping is invertible, and the perception is feasible to be reconstructed from brain activity patterns if the inverse mapping is precisely estimated.

Brain decoding can be distinguished into three categories: identification, classification and reconstruction. The first two have been recorded promising results~\cite{a2,a4,a6,a7} while the last one remains unsolved. This is because accurate reconstruction of perceived images requires both high-level semantic knowledge and low-level visual details. It is hard to extract all this information from noisy high-dimensional brain activity signals (\emph{e.g.}, the fMRI data). In addition, the modality gap commonly exists between cognitive signals and visual stimuli, making the mapping relationship even harder to estimate. Although there have been several attempts at reconstructing the visual stimuli from brain responses~\cite{a11,a12}, they generally yield blurry, cluttered, and low-quality results.

Driven by the success of latent models, adversarial learning and knowledge distillation, we design a novel framework, called Dual-Variational Autoencoder/ Generative Adversarial Network ({\scshape{D-Vae/Gan}}), to learn the mapping from fMRI signals to their corresponding visual stimuli (images). Our goal is to learn an encoder which can map the brain signal to a low-dimensional representation that preserves important visually-relevant information, as well as a powerful decoder that is able to recover this latent representation back to the corresponding visual stimulus. To achieve this goal, our {\scshape{D-Vae/Gan}} first generates low-dimensional latent features for both fMRI signals and perceived images (visual stimuli) with a Dual VAE-Based Encoding Network, then learns to capture visually important information and overcome modality gap between them in this space through a novel GAN-Based inter-modality knowledge distillation method, and finally decodes the learned cognitive latent features back to corresponding visual images using our Adversarial Decoding Network. Intuitively, our solution provides two major advantages. First, by learning lower-dimensional latent representation for the high-dimensional brain signal, the approach is able to filter out the majority of noise and clutters, and thus produces a more compressed representation. Second, by adapting cognitive features to visual features in the lower-dimensional latent space, the approach is able to learn visually important features and overcome the modality gap more effectively.

To sum up, our {\bf contributions} are as follows: {\bf (i)} We introduce a new framework, namely D-VAE/GAN, established by combining our proposed Dual VAE-Based Encoding Network with the Adversarial Decoding Network, to learn a powerful encoder-decoder model for mapping fMRI signals to the corresponding visual stimuli. {\bf (ii)} We propose to leverage \emph{free-of-cost} guidance knowledge from the visual domain for guiding the latent representation learning, without additional annotations. {\bf (iii)} We present a novel three-stage training approach, in which a GAN-Based inter-modality knowledge distillation method is introduced to make the learned cognitive latent features to capture more visually-relevant knowledge by forcing the cognitive encoder to mimic its paired visual encoder. {\bf (iv)} Our method has achieved quite effective reconstruction results on four public fMRI datasets and made an encouraging breakthrough in visual stimuli reconstruction.


\section{Related Works}

Traditionally, machine learning methods play significant roles in the fMRI-Based brain decoding tasks. Miyawaki \emph{et al.}, for the first time, proposed the spares multinomial logistic regression (SMLR) by using multi-voxel patterns of fMRI signals and multi-scale visual representation to reconstruct the lower-order information~\cite{a9}. Schoenmakers \emph{et al.} reconstructed handwritten characters using a straightforward linear Gaussian approach~\cite{a21}. Fujiwara \emph{et al.} proposed to build a reconstruction model to automatically estimate image bases by Bayesian canonical correlation analysis (BCCA)~\cite{a22}. However, because the linear hypothesis in the proposed models does not conform to the actual visual encoding-decoding process in human brain, their performance is far from satisfactory.

The recent integration of deep learning into neural decoding has been a very successful endeavor~\cite{a25,a26}. Gerven \emph{et al.} reconstructed handwritten digits using deep belief networks~\cite{a20}. Several proposed deep multi-view representation learning models, such as deep canonically correlated autoencoders (DCCAE)~\cite{a31} and correlational neural networks (CorrNet)~\cite{a32}, had the ability to learn deep correlational representations, and thus were able to reconstruct each view respectively. However, directly applying the nonlinear maps of DCCAE and CorrNet to limited noisy brain activities is prone to overfitting. A recent neural decoding method was based on multivariate linear regression and deconvolutional neural network (De-CNN)~\cite{a24}. After that, Du \emph{et al.} introduced Bayesian deep learning to study visual image reconstruction~\cite{a13}, which can be viewed as a nonlinear extension of the linear method BCCA. In addition, Güçlütürk \emph{et al.} reconstructed perceived faces with a deep adversarial neural decoding (DAND) model by combining probabilistic inference with deep learning~\cite{a26}. Seeliger \emph{et al.} trained a deep convolutional generative adversarial network to generate gray scale photos~\cite{a27}. More recently, Shen \emph{et al.} took advantage of deep neural networks (DNNs) to reconstruct color photos~\cite{b1,b2}. However, the generated visual stimuli by all existing deep learning-based methods are still rather far from the real ones.

\section{Methodology}
In this section, we first formally define the brain-driven visual stimulus reconstruction problem, give an overview of our framework ({\scshape{D-Vae/Gan}}), and then introduce each component of our model in details. Finally, we show how to train our {\scshape{D-Vae/Gan}} with a novel three-stage training method.

\begin{figure*}[t]
	\begin{center}
		\includegraphics[width=0.77\linewidth]{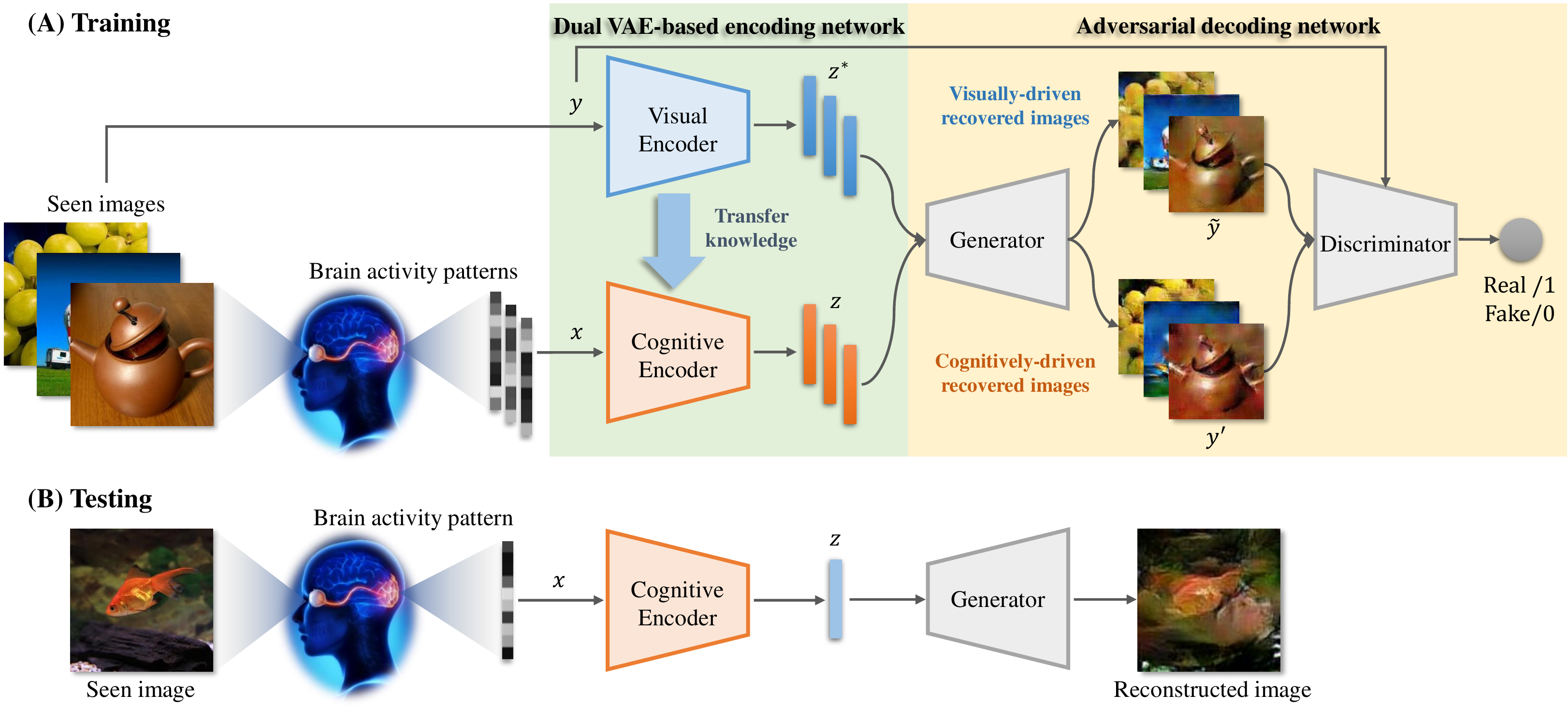}
	\end{center}
	\caption{Overview of our D-VAE/GAN framework, which consists of a Dual VAE-Based Encoding Network and a Adversarial Decoding Network. (A) During the training phase, we have four major components: Cognitive Enconder  ($E_{{\rm{Cog}}}$) , Visual Encoder ($E_{{\rm{Vis}}}$), Generator ($G$) and Discriminator ($D$). (B) During the test process, we only use the trained $E_{{\rm{Cog}}}$ and $G$ to generate the visual stimulus images.}
	\label{fig1}
\end{figure*}

\subsection{Problem Definition and Notations}
In this work, we study the visual stimulus (image) reconstruction problem. Assume $\mathcal{S} = \{(\mathbf{x}_i, \mathbf{y}_i) | \mathbf{x}_i \in \mathcal{M},  \mathbf{y}_i \in \mathcal{I_{M}}\}_{i=1}^n$ are given as training data, where $\mathcal{M}$ is the set of fMRI data and $\mathcal{I_{M}}$ is the corresponding visual stimulus images. The goal of this task is to learn a mapping function $f: \mathcal{M} \rightarrow \mathcal{I_{M}}$, so that for any novel input from test set $\mathbf{x}_t \in \mathcal{T}$, its corresponding visual stimulus can be generated through $\mathbf{y}_t = {f}(\mathbf{x}_t) $. However, directly learning the mapping function $f$ is rather challenging due to the significant heterogeneity gap. To alleviate this problem, we formulate the mapping $f$ as a two-stage process:  1) latent feature learning $f_1: \mathcal{M} \rightarrow z_{\mathcal{M}}$ which transforms real brain signals to latent representations; and 2) visual image reconstruction $f_2: z_{\mathcal{M}} \rightarrow {\mathcal{I_M}}$ that reconstructs stimulus images based on their latent features. Importantly, in the training phase, we also learn the latent representations for real visual stimulus, \emph{i.e.,} $\{\mathbf{z}^{*}_i|\mathbf{z}^{*}_i= f^{*}_1(\mathbf{y}_i)\in z_{\mathcal{M_I}}\}_{i=1}^n$ where $f^{*}_1: \mathcal{I_M} \rightarrow z_{\mathcal{I_M}}$ denotes the mapping function, and treat them as exemplars for guiding the learning of latent representations of brain signals (fMRI data) $\{\mathbf{z}_i|\mathbf{z}_i= f_1(\mathbf{y}_i)\in z_{\mathcal{M}}\}_{i=1}^n$. The visual latent representations $z_{\mathcal{I_M}}$ and cognitive latent representations $z_{\mathcal{I}}$ share the same reconstruction function so as to enforce them to be close in latent space , \emph{i.e.}, $f_2: z_{\mathcal{M}} \cup z_{\mathcal{I_M}} \rightarrow {\mathcal{I_M}}$. 

\subsection{Overview of Framework}
The overall framework of our {\scshape{D-Vae/Gan}} is illustrated in Figure~\ref{fig1}. There are three major components in our framework: Cognitive Enconder  ($E_{{\rm{Cog}}}$) , Visual Encoder ($E_{{\rm{Vis}}}$), Generator ($G$) and Discriminator($D$). The core component of {\scshape{D-Vae/Gan}} is the Cognitive Encoder network that can transform the high-dimensional and noisy brain signals (fMRI) to low-dimensional latent representations. Along with $E_{{\rm{Cog}}}$, we have a Visual Encoder network that maps the visual stimuli into the latent representations. Intuitively, $E_{{\rm{Vis}}}$ is more capable of capturing all visually-related information during the encoding process, since it directly takes visual stimuli as inputs. Therefore the generated latent features by $E_{{\rm{Vis}}}$ can be considered as the \emph{free-of-cost} guidance knowledge for $E_{{\rm{Cog}}}$ to learn the latent representations. From the standpoint of knowledge distillation, we can treat $E_{{\rm{Vis}}}$ as a \emph{teacher} network, and train $E_{{\rm{Cog}}}$ (\emph{student} network) by transferring knowledge from it. The Generator network in our framework is used to rebuild the visual stimuli based on the latent representations. Note that, during the training phase, the $E_{{\rm{Cog}}}$ and $E_{{\rm{Vis}}}$ share the same $G$ in our framework, which forces $E_{{\rm{Cog}}}$ (student) to mimic $E_{{\rm{Vis}}}$ (teacher). Finally, we have a Discriminator network that acts as a classifier to distinguish real visual stimuli from fake stimuli produced by the generator. Note that the visual stimuli are used to guide cognitive latent feature learning and supervise the generation of reconstructed images \emph{only} in the training stage. In the test stage, the \emph{only} inputs are the cognitive signals (fMRI).

\subsection{Dual VAE-Based Encoding Network}
In the training phase, we include two encoders in our framework, Cognitive Enconder  ($E_{{\rm{Cog}}}$) and Visual Encoder ($E_{{\rm{Vis}}}$), which form a dual encoding network architecture. Generally, both $E_{{\rm{Cog}}}$ and $E_{{\rm{Vis}}}$ are feed-forward neural networks and aim to encode a data sample $\mathbf{x}_i$ (or $\mathbf{y}_i$) to the corresponding latent representation $\mathbf{z} \in z_{\mathcal{M}}$ (or $\mathbf{z}^{*} \in z_{\mathcal{M_I}})$. Ideally, the learned latent representation $\mathbf{z}$ (or $\mathbf{z}^*$) should capture important visually-related information (including the texture, color, location, stylistic properties, \emph{etc}) such that the generator (G) is able to reconstruct the visual stimuli from these latent features.

\subsubsection{Cognitive Encoder ($E_{{\rm{Cog}}}$).}
To learn a robust cognitive latent representation $\mathbf{z}$ for each input fMRI $\mathbf{x}$, we resort to the Variational Autoencoder (VAE)-based encoder, \emph{i.e.}, \emph{probabilistic} encoder, to model the mapping $f_1: \mathcal{M} \rightarrow z_{\mathcal{M}}$. Specifically, $E_{{\rm{Cog}}}$ is a lightweight neural network which is composed of a fully connected layer and an output layer. Given a cognitive pattern $\mathbf{x}$, $E_{{\rm{Cog}}}$ produces a distribution over the possible values of the cognitive latent representation $\mathbf{z}$ from which the corresponding visual stimulus $\mathbf{y} \in \mathcal{I_{M}}$ could be generated. Following ~\cite{kingma2013auto}, the cognitive latent feature $\mathbf{z}$ can be computed as follows:
\begin{equation}
{\mathbf{z} \sim  E_{{\rm{Cog}}}(\mathbf{x}; \theta) = q_{\theta}\left( {\mathbf{z}\left| \mathbf{x} \right.} \right),}
\end{equation}
\noindent where $q_{\theta}\left( {\mathbf{z}\left| \mathbf{x} \right.} \right)$ is a Gaussian distribution whose mean $\mathbf{\mu}$ and diagonal covariance $\Sigma$ are the output of $E_{{\rm{Cog}}}$ parameterized by $\theta$. 

Here, we regularize the distribution $q_{\theta}\left( {\mathbf{z}\left| \mathbf{x} \right.} \right)$ to be a simple Gaussian distribution based on the Kullback-Leibler (KL) divergence between $q_{\theta}\left( {\mathbf{z}\left| \mathbf{x} \right.} \right)$ and the prior distribution $p({\mathbf{z}}) = \mathcal{N}(0,\mathbf{I})$,
\begin{equation}
\begin{aligned}
\mathcal{L}_{{\rm{prior}-c}} (\mathbf{x}, \mathbf{z}; \theta) &=D_{KL} (q_{\theta}\left( {\mathbf{z}\left| \mathbf{x} \right.} \right) || p({\mathbf{z}}))\\
&= D_{KL} [\mathcal{N}(\mathbf{\mu},\Sigma) || \mathcal{N}(0,\mathbf{I})],
\end{aligned}
\end{equation}
\noindent where $D_{KL}(\cdot)$ means the KL divergence, and $\mathbf{I}$ is the identity matrix.

\subsubsection{Visual Encoder ($E_{{\rm{Vis}}}$).} Our visual encoder is also a VAE-based encoder, which is used to model the mapping $f^{*}_1: \mathcal{I_M} \rightarrow z_{\mathcal{I_M}}$. But differently, $E_{{\rm{Vis}}}$ takes an image $\mathbf{y} \in \mathcal{I_{M}}$ rather than an fMRI sequence as input, which makes it different from $E_{{\rm{Cog}}}$ in network architecture. Specifically, compared with $E_{{\rm{Cog}}}$, $E_{{\rm{Vis}}}$ has three additional convolutional layers and one reshape layer on top of the fully connected layer so as to transform the imput image to the latent representation. Mathematically, the visual latent feature $\mathbf{z}^*$ is given as:
\begin{equation}
{\mathbf{z}^* \sim  E_{{\rm{Vis}}}(\mathbf{y}; \phi) = q_{\phi}\left( {\mathbf{z}^*\left| \mathbf{y} \right.} \right),}
\end{equation}
\noindent where $\mathbf{z}^*$ is drawn from the posterior $q_{\phi}\left( {\mathbf{z}^*\left| \mathbf{x} \right.} \right)$ which is of Gaussian learned by $E_{{\rm{Vis}}}$ with learnable parameters $\phi$. 

Similarly, the mean $\mathbf{\mu}^*$ and diagonal covariance $\Sigma^*$ of the distribution $q_{\phi}\left( {\mathbf{z}^*\left| \mathbf{y} \right.} \right)$ is regularized by
\begin{equation}
\begin{aligned}
\mathcal{L}_{\rm{prior}-{v}} (\mathbf{y}, \mathbf{z}^*; \phi) &=D_{KL} (q_{\phi}\left( {\mathbf{z}^*\left| \mathbf{y} \right.} \right) || \hat{p}({\mathbf{z^*}}))\\
&= D_{KL} [\mathcal{N}(\mathbf{\mu}^*,\Sigma^*) || \mathcal{N}(0,\mathbf{I})],
\end{aligned}
\end{equation}
\noindent where $\hat{p}({\mathbf{z}^*})$ denotes the prior distribution, and $\hat{p}({\mathbf{z}^*}) = \mathcal{N}(0,\mathbf{I})$.

\subsection{Adversarial Decoding Network}
Given the learned latent representations, including cognitive latent feature ${\mathbf{z}}$ and visual latent feature $\mathbf{z}^*$, the generator tries to reconstruct the corresponding visual stimulus $\mathbf{y}$. Here, the VAE decoder and the GAN generator are integrated by letting them share parameters. Through the above combination, we can take advantages of both VAE and GAN. On one hand, the adversarial loss computed from the discriminator makes generated visual stimulus more realistic. On the other hand, the optimization for VAE-Based decoder become quite stable, and thus avoids the ``collapse'' problem. We name such combined decoder network as Adversarial Decoding Network which includes a VAE-Based Decoder/ Generator ($G$) and a GAN discriminator ($D$), and use it to model the mapping $f_2: z_{\mathcal{M}} \cup z_{\mathcal{I_M}} \rightarrow {\mathcal{I_M}}$. 

\subsubsection{VAE-Based Decoder/ Generator ($G$).} 
Our VAE-Based decoder $G$ is symmetric to the visual encoder $E_{{\rm{Vis}}}$, and it has a fully connected layer, a reshape layer, and three de-convolutional layers. For clarity, we denote the generator ${p_\eta}(\tilde{\mathbf{y}} | {\mathbf{z}'})$ with learnable parameters $\eta$. For any latent feature $\mathbf{z}' \in z_{\mathcal{M}} \cup z_{\mathcal{M_I}}$, $G$ estimates the  corresponding visual stimulus $\tilde{\mathbf{y}}$, and thus,
\begin{equation}
\tilde{\mathbf{y}} =  G(\mathbf{z'}; \eta) ={p_\eta}(\tilde{\mathbf{y}} | {\mathbf{z}'}),
\end{equation}

\noindent where $\eta$ is the learnable network parameters of $G$. Our goal is to make $\tilde{\mathbf{y}}$ similar to $\mathbf{y}$ as much as possible. We update the network parameters $\eta$ of the VAE-Based decoder $G$ jointly with the given encoder $E_{{\rm{Cog}}}$ or $E_{{\rm{Vis}}}$. Take the combination of $E_{{\rm{Vis}}}$ and $G$ for example, the parameters $\eta$ and $\phi$ are jointly updated by minimizing the loss:
\begin{equation}
\begin{aligned}
\mathcal{L_{V}}& (\mathbf{y}, \mathbf{z}^*, \tilde{\mathbf{y}}; \eta, \phi)= - \mathbb{E}_{q_{\phi}(\mathbf{z}^*|{\mathbf{y}})} [\log\cfrac{{p_\eta}({\tilde{\mathbf{y}}} | {\mathbf{z}^*}) p({\mathbf{z}^*})}{q_{\phi}(\mathbf{z}^* | {{\mathbf{y}}})}]\\
&=- \mathbb{E}_{q_{\phi}(\mathbf{z}^*|{\mathbf{y}})}[{\log p_\eta}({\tilde{\mathbf{y}}} | {\mathbf{z}^*})] + D_{KL} (q_{\phi}\left( {\mathbf{z}^*\left| \mathbf{y} \right.} \right) || \hat{p}({\mathbf{z^*}}))\\
&= \mathcal{L}_{\rm{rec}} + \mathcal{L}_{\rm{prior-v}},
\end{aligned}
\label{eq6}
\end{equation}
\noindent where $\mathcal{L}_{\rm{rec}} = - \mathbb{E}_{q_{\phi}(\mathbf{z}^*|{\mathbf{y}})}[{\log p_\eta}({\mathbf{\tilde{y}}} | {\mathbf{z}^*})]$ denotes the reconstruction loss, and $\mathcal{L}_{\rm{prior-v}}$  means the KL divergence~\cite{a38} between the encoder distribution and a known prior. 

\begin{figure}[pt]
	\begin{center}
		\includegraphics[width=0.99\linewidth]{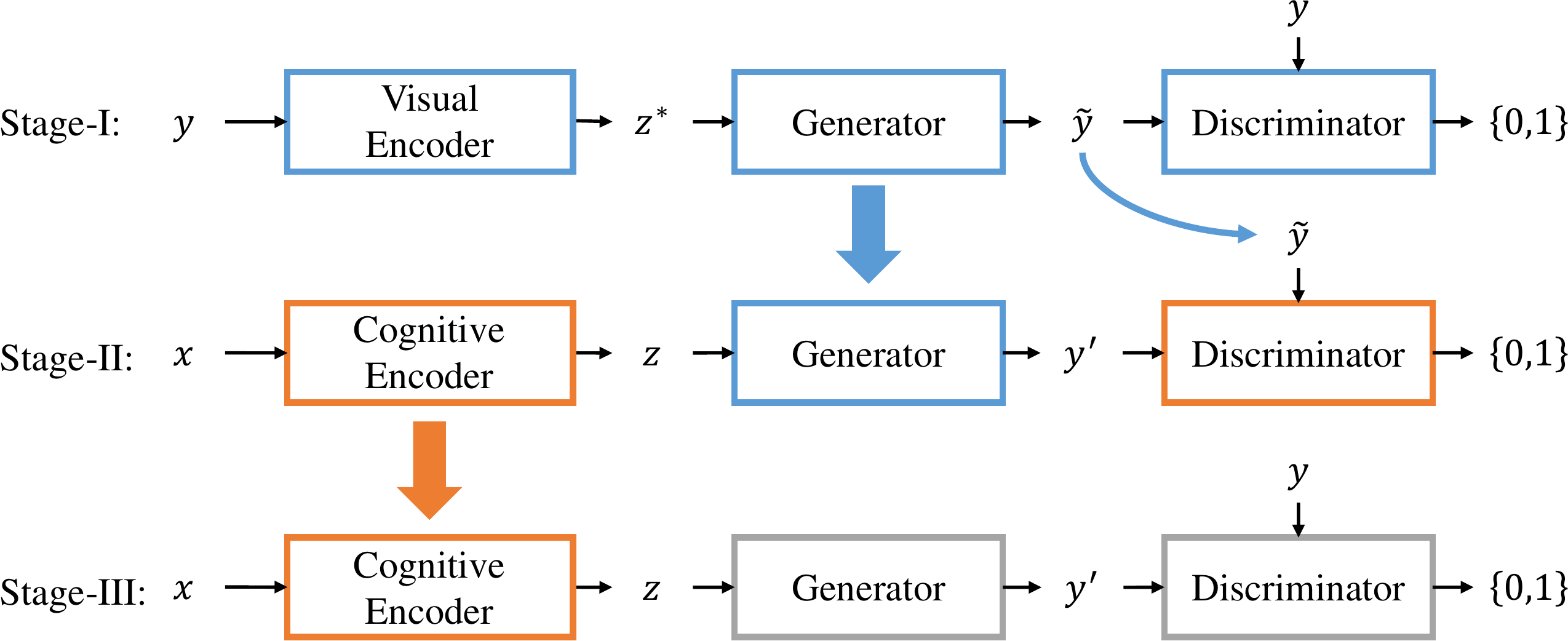}
	\end{center}
	\caption{Overview of our three-stage training method. Stage-I: training $E_{{\rm{Vis}}}$, $G$ and $D$. Stage-II: keeping $G$ fixed, and $\tilde{\mathbf{y}}$ produced in Stage-I is defined as the real for training $E_{{\rm{Cog}}}$; Stage-III: keeping $E_{{\rm{Cog}}}$ fixed and treating $\mathbf{y}$ as the real for updating the $G$ and $D$.}
	\label{fig2}
\end{figure}

\subsubsection{GAN Discriminator ($D$).}
Another important component in our adversarial decoding network is the GAN discriminator ($D$), which assigns ${\mathbf{c}} \in [0, 1]$ for the input $\mathbf{y}$ to measure the probability $\mathbf{q}$ that ${\mathbf{y}}$ is a real visual stimulus. Specifically, a standard discriminator $D$ with adversarial training on its parameter $\tau$ is used to further improve the quality of $\tilde{\mathbf{y}}$. The adversarial training loss for discriminator is formulated as,
\begin{equation}
\mathcal{L}_{{\rm{gan}}} ({\mathbf{y}}, \tilde{\mathbf{y}}, {\mathbf{c}}; \tau) = \log(D({\mathbf{y}}, \tau)) + \log((1-\tilde{\mathbf{y}}), \tau)),
\end{equation}
\noindent where $\mathbf{y}$ means the real visual stimulus. 

In addition, we can further replace $\mathcal{L}_{\rm{rec}}$ with a reconstruction loss expressed by the GAN discriminator. Here, we still take the combination of $E_{{\rm{Vis}}}$ and $G$ for example. Following~\cite{Zheng_2019_CVPR,a18}, we employ a Gaussian observation model for $D_l({\mathbf{y}}, \tau)$ with $D_l({\tilde{\mathbf{y}}}, \tau)$ and identity covariance:
\begin{equation}
p_{\eta}(D_l({\mathbf{y}}, \tau) | \mathbf{z}^*) = \mathcal{N}(D_l({ {\mathbf{y}}}, \tau) | D_l({\tilde{\mathbf{y}}}, \tau), \mathbf{I}),
\end{equation}
\noindent where $D_l(\cdot)$ denotes the hidden representation of $l_{th}$ layer of the discriminator. Therefore, $\mathcal{L}_{\rm{rec}}$ in Eq.~\ref{eq6} can be replaced by
\begin{equation}
\begin{aligned}
\mathcal{L}^{\rm{gan}}_{\rm{rec{-}v}} (\mathbf{y}, \mathbf{z}^*, \tilde{\mathbf{y}}; \eta, \phi, \tau) = - \mathbb{E}_{q_{\phi}(\mathbf{z}^*|{\mathbf{y}})}[\log p_{\eta}(D_l({\mathbf{y}}, \tau) | \mathbf{z}^*)].
\end{aligned}
\end{equation}

\subsection{Three-Stage Training Method}
As illustrated in Figure \ref{fig2}, three-stage training method is used to train our model in a gradual fashion.

\subsubsection{Stage-I: Visual Latent Fearure Learning.}

In the first stage, the visual stimuli $\mathcal{I_{M}}$ are used as input to jointly train the visual encoder $E_{{\rm{Vis}}}$, VAE-Based generator $G$ and GAN discriminator $D$. In this stage, we perform an {\bfseries intra-modality reconstruction task} (within the same modality) for learning the visual latent representations. Specifically, $E_{{\rm{Vis}}}$ is optimized to encode the visual stimulus $\mathbf{y} \in \mathcal{I_{M}}$ to the visual latent representation $\mathbf{z}^*$, and then $G$ and $D$ are trained to reconstruct $\mathbf{y}$ given $\mathbf{z}^*$. In this stage, we train network  parameters $\phi$ of $E_{{\rm{Vis}}}$, $\eta$ of $G$ and $\tau$ of $D$ with a triple criterion:
\begin{equation}
\begin{aligned}
&\mathcal{L_V} (\mathbf{y}, \mathbf{z}^*, \tilde{\mathbf{y}}, {\mathbf{c}}; \phi, \eta, \tau) =
\mathcal{L}^{\rm{gan}}_{\rm{rec{-}v}} (\mathbf{y}, \mathbf{z}^*, \tilde{\mathbf{y}}; \eta, \phi, \tau)\\
&+\mathcal{L}_{\rm{prior}-{v}} (\mathbf{y}, \mathbf{z}^*; \phi) + \mathcal{L}_{{\rm{gan}}} ({\mathbf{y}}, \tilde{\mathbf{y}}, {\mathbf{c}}; \tau).
\end{aligned}
\label{eq10}
\end{equation}

\subsubsection{Stage-II: Cognitive Latent Fearure Learning.} 

In the Stage-II, we perform an {\bf inter-modality knowledge distillation} task that forces the cognitive encoder to mimic the visual encoder, so as to capture more visually important features in the cognitive latent space. Intuitively, it makes sense that $E_{{\rm{Vis}}}$ is more capable of capturing visually important features for the visual stimuli reconstruction since it learns latent representation within the same modality. Therefore, we treat $E_{{\rm{Vis}}}$ as a \emph{teacher} network and $E_{{\rm{Cog}}}$ as a \emph{student} network, and train $E_{{\rm{Cog}}}$ to mimic $E_{{\rm{Vis}}}$ so as to better capture visually-related information from noise brain signals.

Aiming to achieve this goal, we fix the generator $G$ and use the images/stimuli $\{\tilde{\mathbf{y}_i}\}_{i=1}^n$ produced by the teacher network $E_{{\rm{Vis}}}$ as the real to the discriminator $D$ for guiding the cognitive latent feature learning of $E_{{\rm{Cog}}}$. Given the fMRI data $\{\mathbf{x}_i\}_{i=1}^n$, we train the network parameters $\theta$ for the cognitive encoder and update $\tau$ of $D$ to encourage the generated visual stimulus ${\mathbf{y}'}$ from the cognitive latent feature $\mathbf{z}$ to be similar with $\tilde{\mathbf{y}}$ which generated from the visual latent feature $\mathbf{z}^*$ through the following loss function:
\begin{equation}
\begin{aligned}
&\mathcal{L_C} (\mathbf{x}, \mathbf{z}, \tilde{\mathbf{y}}, \mathbf{y'}, {\mathbf{c}}; \theta, \tau) = \mathcal{L}^{\rm{gan}}_{\rm{rec{-}c}}(\mathbf{x}, \mathbf{z}, {\mathbf{y}'}; \theta, \tau)\\
&+\mathcal{L}_{{\rm{prior}-c}} (\mathbf{x}, \mathbf{z}; \theta) + \mathcal{L}_{{\rm{gan}}} (\tilde{\mathbf{y}}, {\mathbf{y'}}, {\mathbf{c}}; \tau).
\end{aligned}
\label{eq11}
\end{equation}

\begin{figure}[pt]
	\begin{center}
		\includegraphics[width=0.88\linewidth]{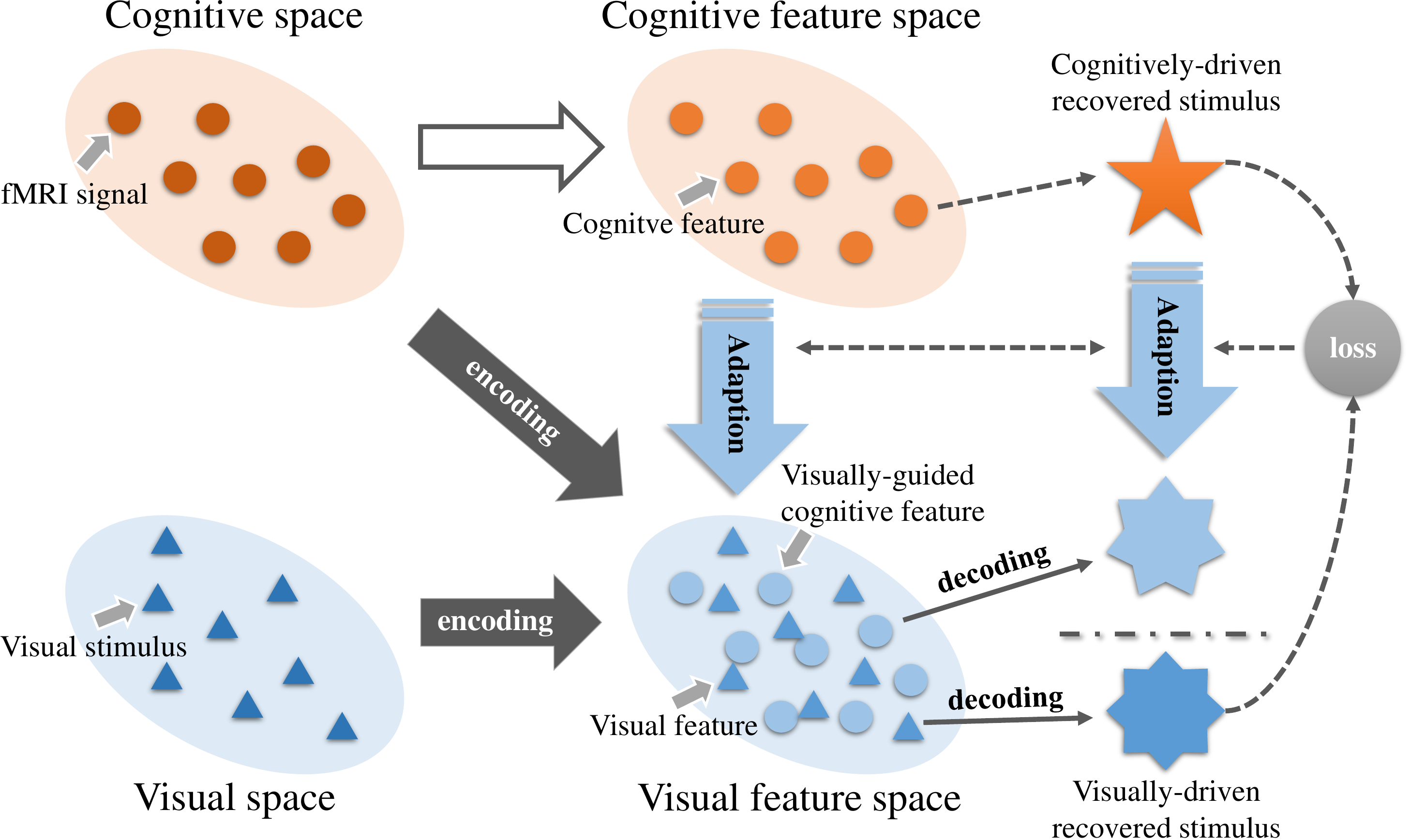}
	\end{center}
	\caption{The illustration of GAN-Based cross-modality knowledge distillation. The knowledge learned by the visual encoder $E_{{\rm{Vis}}}$ (teacher) is gradually transferred to the cognitive encoder $E_{{\rm{Cog}}}$ (student) through the adversarial loss.}
	\label{fig3}
\end{figure}

As illustrated in Figure \ref{fig3}, the cognitive encoder $E_{{\rm{Cog}}}$ is forced to mimic the visual encoder $E_{{\rm{Vis}}}$ by minimizing Eq.~\ref{eq11}, and generates the latent feature $\mathbf{z}$ which is similar to $z^{*}$. Our GAN-Based inter-modality knowledge distillation method is a generalization of  the distillation idea from~\cite{hinton2015distilling}, and enables distillation of knowledge from the other modality. 

\subsubsection{Stage-III: Visual Stimuli Recovering.} 

Finally, in the third stage, we perform a {\bf fine-tuning} task to slightly update the network parameters $\eta$ of the generator $G$ such that it can become more capable of recovering the visual stimuli when given cognitive latent feature $\mathbf{z}$. In this stage, we fix parameters of the cognitive encoder $E_{{\rm{Cog}}}$ and use the visual stimulus $\mathbf{y} \in \mathcal{I_{M}}$ as the real to $D$ for fine-tuning $\eta$. The loss function is given as:
\begin{equation}
\mathcal{L'_C} ({\mathbf{z}}, {\mathbf{y}}, {\mathbf{c}}; \eta, \tau) = \mathcal{L}_{{\rm{gan}}} ({\mathbf{z}}, {\mathbf{y}}, {\mathbf{c}}; \eta, \tau).
\end{equation}

Our three-stage training method is carefully designed to mitigate the gap between source and target domains in the training process. Our experiments in the next section demonstrate that the proposed three-stage training method is effective at learning good cognitive latent features from fMRI signals, and these learned latent features can be successfully recovered back to visual stimuli.

\section{Experiments and Results}

In this section, we evaluate the proposed {\scshape{D-Vae/Gan}} on four benchmark datasets. Following the convention of existing work~\cite{b3,b1,b2}, both objective and subjective assessment methods are used for evaluation. 

\subsection{Dataset.} 

 \begin{table}[pt]
	\begin{center}
		\caption{The details of the four datasets used in our experiments. ROIs indicate the related visual regions of interest in brain.}\smallskip
		\renewcommand\arraystretch{1.1}
		\centering
		\resizebox{1\columnwidth}{!}{
			\smallskip\begin{tabular}{c|c|c|c|c|c}
				
				{\bf Dataset} &{\bf Instance}&{\bf Resolution}&{\bf Voxel}&{\bf ROIs}&{\bf Training}\\
				\hline
				Dataset1&  2040 & $10\times 10$ & 5438 & V1 &1320 \\
				
				Dataset2& 100 & $28\times 28$ & 3092 & V1,V2,V3 & 90\\
				
				Dataset3& 360 & $56\times 56$ & 2420 & V1,V2 & 288\\
				
				Dataset4& 1250 &  $500\times 500$ & 4466 & V1, V2, V3, V4, LOC, FFA, PPA & 1200\\
				
			\end{tabular}
		}
	\end{center}
	\label{tab:1}
\end{table}

We evaluated the reconstruction quality on four public datasets. The information about the four datasets is summarized in Table~\ref{tab:1}.

\begin{table*}[pt]
	\centering
	\caption{Reconstruction performance evaluated by standard image similarity metrics (PCC and SSIM) on the three datasets.}\label{tab2}
	\resizebox{0.85\textwidth}{!}{
		\begin{tabular}{l|c|c|c|c|c|c}
			\hline
			
			\multirow{2}{*}{Methods} &\multicolumn{2}{c|}{Dataset1} &\multicolumn{2}{c|}{Dataset2} &\multicolumn{2}{c}{Dataset3}\\
			\cline{2-7}
			&PCC &SSIM &PPC &SSIM &PPC &SSIM\\
			\hline
			\hline
			\quad{\small SMLR~\cite{a9}} &.609$\pm$.151 &.237$\pm$.105 &.767$\pm$.033&.466$\pm$.030 &.481$\pm$.096 &.191$\pm$.043\\
			
			\quad{\small BCCA~\cite{a22}} &.438$\pm$.215 &.181$\pm$.066 &.411$\pm$.157&.192$\pm$.035 &.348$\pm$.138 &.058$\pm$.042\\
			
			\quad{\small DCCAE-A~\cite{a31}} &.455$\pm$.113 &.166$\pm$.025 &.548$\pm$.044&.358$\pm$.097 &.354$\pm$.167 &.186$\pm$.234\\
			
			\quad{\small DCCAE-S~\cite{a31}} &.401$\pm$.100 &.175$\pm$.011&.511$\pm$.057&.552$\pm$.088 &.351$\pm$.153 &.179$\pm$.117\\

			\quad{\small De-CNN~\cite{a24}} &.469$\pm$.149 &.224$\pm$.129&.799$\pm$.062&.613$\pm$.043 &.470$\pm$.149 &.322$\pm$.118\\
			
			\quad{\small DGMM~\cite{a13}} &.611$\pm$.183 &.268$\pm$.106&.803$\pm$.063&.645$\pm$.054&.498$\pm$.193 &.340$\pm$.051\\
			
						\quad{\small Denoiser GAN~\cite{a27}} & - & - &.531$\pm$.049& .529$\pm$.043&.319$\pm$.032 &.465$\pm$.031\\
			
			\quad{\small DGMM+~\cite{du2018reconstructing}} &.631$\pm$.153 &.278$\pm$.106&.813$\pm$.053&.651$\pm$.044&.502$\pm$.193 &.360$\pm$.050\\
			\hline
			
			\quad{\small {\scshape{DCGAN}}} &{-.123$\pm$.008} &{-.006$\pm$.006}&{.492$\pm$.030}&{.400$\pm$.016 }&{.023$\pm$.003} &{.002$\pm$.001}\\
			\quad{\small {\scshape{\scshape{Vae/Gan}}}} &{.002$\pm$.009} &{.007$\pm$.005}&{.534$\pm$.043}&{.414$\pm$.017}&{.152$\pm$.026} &{.277$\pm$.010}\\
			\quad{\small {\scshape{\scshape{D-Cnn/Gan}}}} &{.073$\pm$.004} &{.071$\pm$.005}&{.481$\pm$.038}&{.430$\pm$.023}&{.136$\pm$.022} &{.300$\pm$.017}\\
			\hline

			\quad{\small {\scshape{D-Vae/Gan}} (Ours)} &{\bfseries.647$\pm$.001} &{\bfseries.283$\pm$.010}&{\bfseries.837$\pm$.014}&{\bfseries.714$\pm$.014}&{\bfseries.740$\pm$.020} &{\bfseries.587$\pm$.019}\\
			\hline
		\end{tabular}
	}
	\label{tab:2}
\end{table*}

	\begin{itemize}
	\item {\bf  Dataset 1: Geometric Shapes and Alphabetical Letters}. This dataset consists of the contrast-defined patches and contains two independent sessions~\cite{a9}. One is the ``random image session'', in which spatially random patterns are sequentially presented. The other one is the ``figure image session'', where alphabetical letters and simple geometric shapes are sequentially presented. The images from the ``random image session'' are used for training and images from the ``figure image session'' are used to test the performance. The fMRI data are taken from primary visual area V1.

	\item {\bf  Dataset 2: Handwritten Digits}. This dataset contains a hundred of handwritten gray-scale digits with the size of 28$\times$28~\cite{a20}. The images are taken from the training set of the MNIST and the fMRI data are taken from V1--V3. 
	
	\item {\bf  Dataset 3: Handwritten Characters}. This dataset contains 360 gray-scale handwritten characters~\cite{a21} . The images have the size of 56$\times$56 which are taken from~\cite{a33} and the fMRI data of V1--V2 are taken from three subjects.
	
  \item {\bf  Dataset 4: Real Color Photos}. This dataset~\cite{a34} is referred to as \texttt{Generic Object Decoding} dataset. It includes images with high resolutions (500 $\times$ 500), and have 150 object categories from the ImageNet. Following~\cite{a34}, 1,200 images from 150 object categories (8 images from each category) are used for training, and 50 images from 50 object categories (no overlap with the training set) are used for testing. The fMRI data are taken from V1--V4, LOC, FFA and PPA. 
  
\end{itemize}

\subsection{Implementation Details and Evaluation Protocol.}
We implement our {\scshape{D-Vae/Gan}} on Tensorflow, with a 1080Ti GPU$\times$11G RAM card on a single server. During the training process, we use Adam optimizer with $\beta=0.9$. The base learning rate is set to $3\times10^{-4}$, and the $decay_{rate}$ is set to 0.98. During the test, our {\scshape{D-Vae/Gan}} takes an fMRI instance as the input and generates a visual image. Note that all images (visual stimuli) are resized to $100\times 100$ for training and evaluation.

We use four widely-used metrics to evaluate the performance of our {\scshape{D-Vae/Gan}}. First, we use two standard image similarity metrics including Pearson's Correlation Coefficient (PCC) and Structural Similarity Index (SSIM)~\cite{a16}. Then, following~\cite{b3,b1,b2}, we further evaluate the quality of results using both objective and subjective assessment methods. For the objective assessment, we compare the pixel-wise correlation coefficients of the reconstructed image with two candidates, in which one is its corresponding original stimulus and the other is a stimulus image randomly selected from the rest test images of the same image type. For the subjective assessment, we conduct a behavioral experiment with a group of 13 raters (5 females and 8 males, aged between 19 and 41 years). The subjects are presented a reconstructed image by using our {\scshape{D-Vae/Gan}} with two candidate images (designed in the same way as the objective assessment), and then are asked to select the option which appears more similar to the reconstructed image. For clarity, we name the objective assessment as Pix-Com, and the subjective assessment as Hum-Com.

\begin{figure}[h]
	\begin{center}
		\includegraphics[width=1\linewidth]{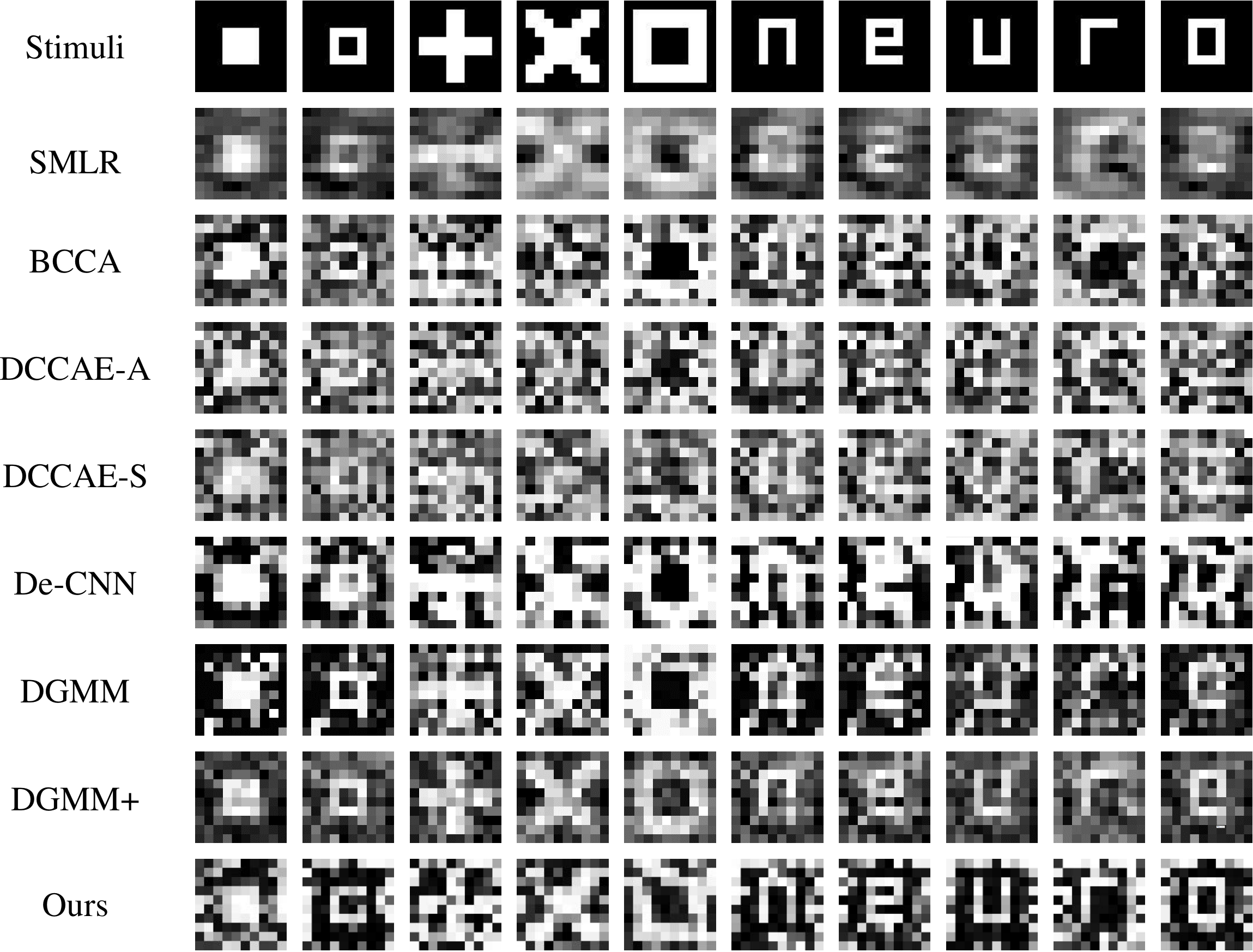}
	\end{center}
	\caption{Reconstructions of geometric shapes and alphabet letters taken from Dataset1.}
	\label{fig4}
\end{figure}

\begin{figure}[h]
	\begin{center}
		\includegraphics[width=1\linewidth]{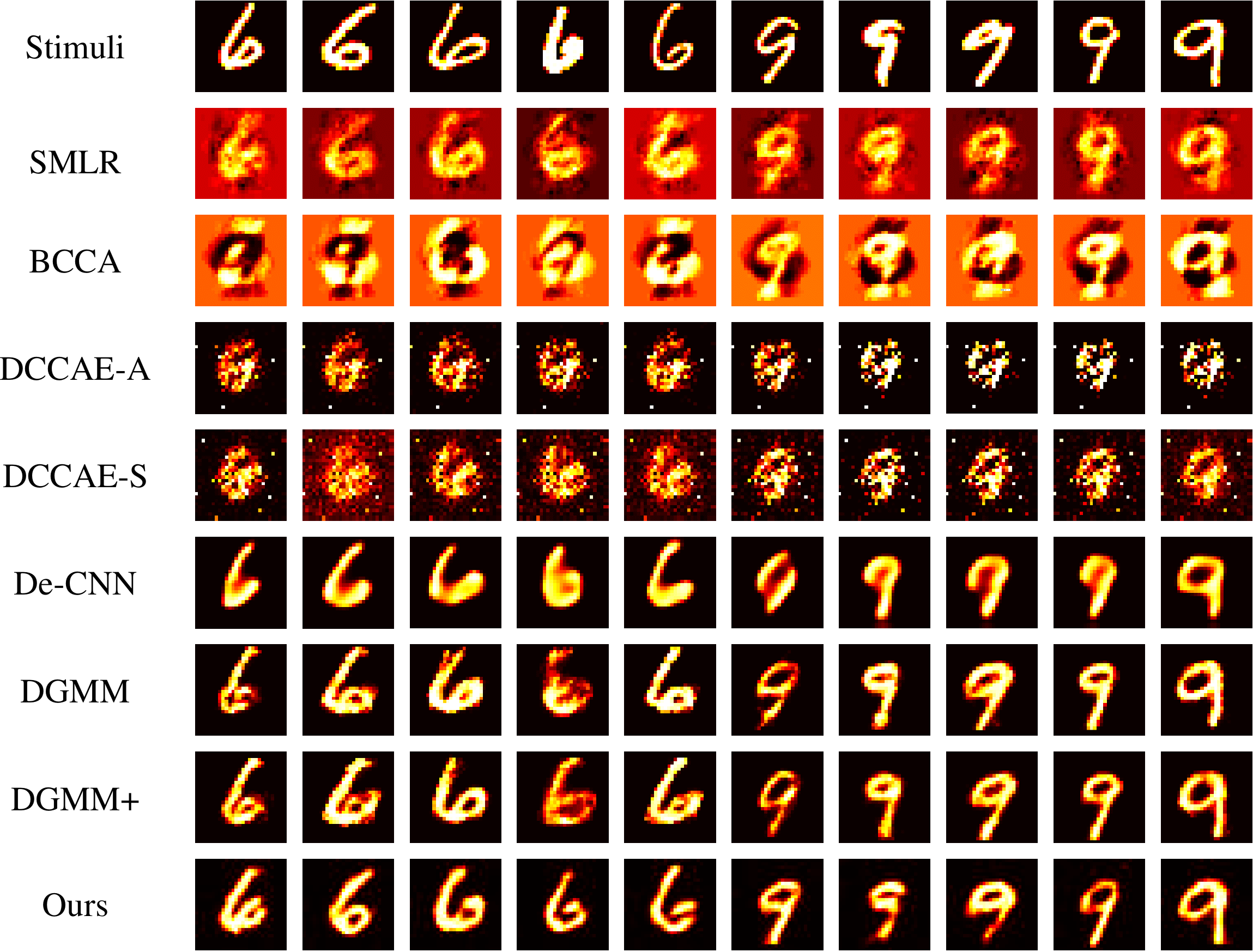}
	\end{center}
	\caption{Reconstructions of 10 distinct handwritten digits taken from Dataset2.}
	\label{fig5}
\end{figure}

\begin{figure*}[h]
	\begin{center}
		\includegraphics[width=1\linewidth]{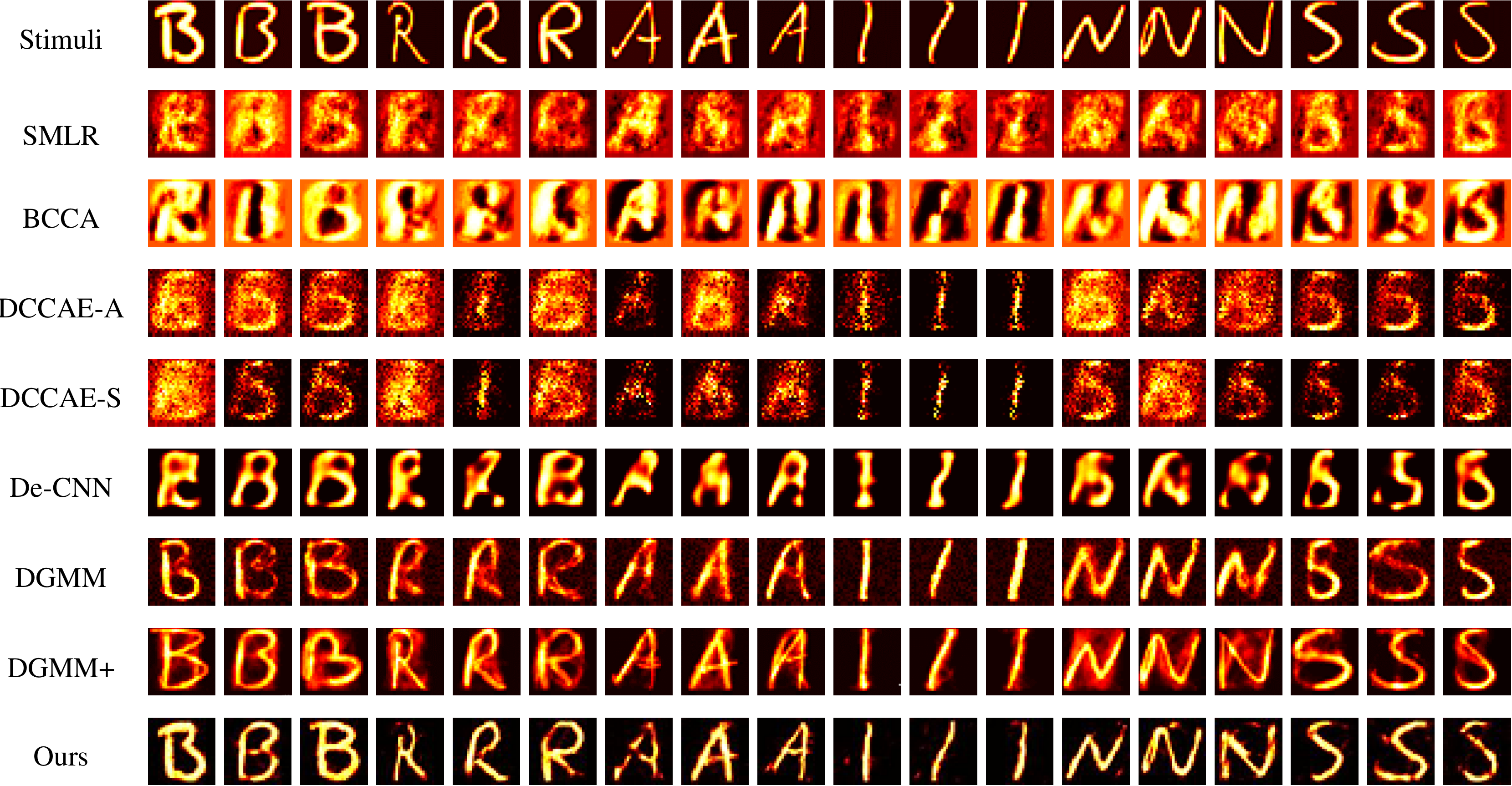}
	\end{center}
	\caption{Reconstructions of 18 distinct handwritten characters taken from Dataset3.}
	\label{fig6}
\end{figure*}

\begin{figure*}[!h]
	\begin{center}
		\includegraphics[width=1\linewidth]{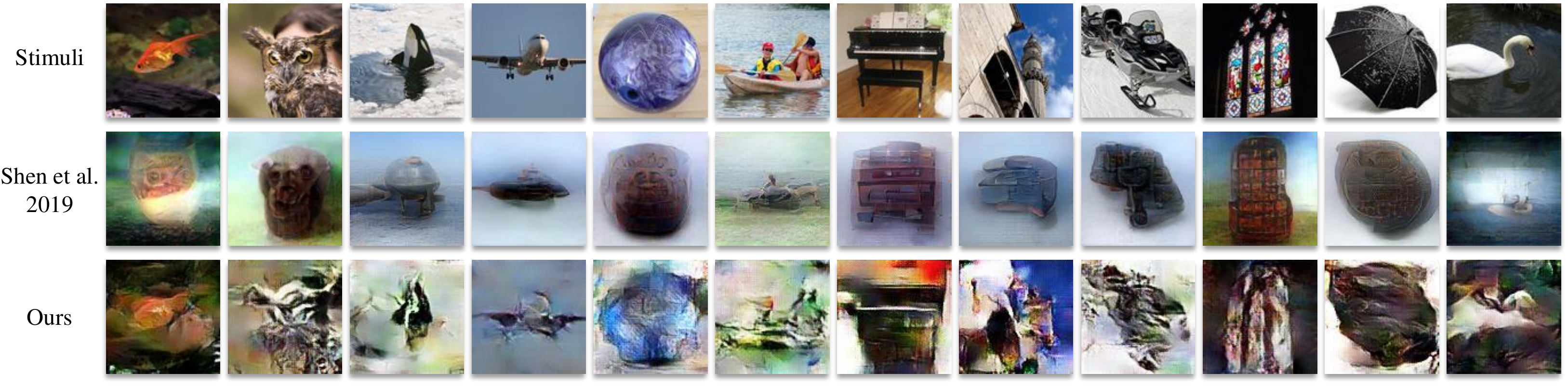}
	\end{center}
	\caption{Reconstructions of real natural photos taken from Dataset4.}
	\label{fig7}
\end{figure*}

\subsection{Comparison with State-of-the-Arts.} 
In this section, we compare our method with other state-of-the-art algorithms. The experiments are conducted on four datasets: {Dataset 1} -- {Dataset 4}. First, we evaluate the performance of our model on some relatively simple visual stimuli ({Dataset 1} -- {Dataset 3}), and then we show results on complex stimuli including images of real scenes ({Dataset 4}). 

As shown in Table~\ref{tab:2}, our {\scshape{D-Vae/Gan}} consistently outperforms the state-of-the-art algorithms on all benchmarks according to PCC and SSIM. We observe that traditional brain decoding methods like SMLR and BCCA only achieve limited performance, which is caused by its linear architecture and spherical covariance assumption. Recent deep learning models achieve significant breakthroughs in the brain signal reconstruct. Compared with recent deep learning-based methods~\cite{a13,a24,a27,du2018reconstructing}, our {\scshape{D-Vae/Gan}} can generate visual stimuli which are more consistent with the perceived images (see in Figure 3, 4, and 5 of the supplemental material). This is because our framework and the novel three-stage training method largely bridge the gap between multi-modal signals, and thus the learned latent representation can capture important visually relevant knowledge which is significant in the recovering process. Some sample results of different methods on {Dataset 1} -- {Dataset 3} are given in Figure ~\ref{fig4}, ~\ref{fig5}, and ~\ref{fig6}.

\begin{figure}
	\begin{center}
		\includegraphics[width=0.71\linewidth]{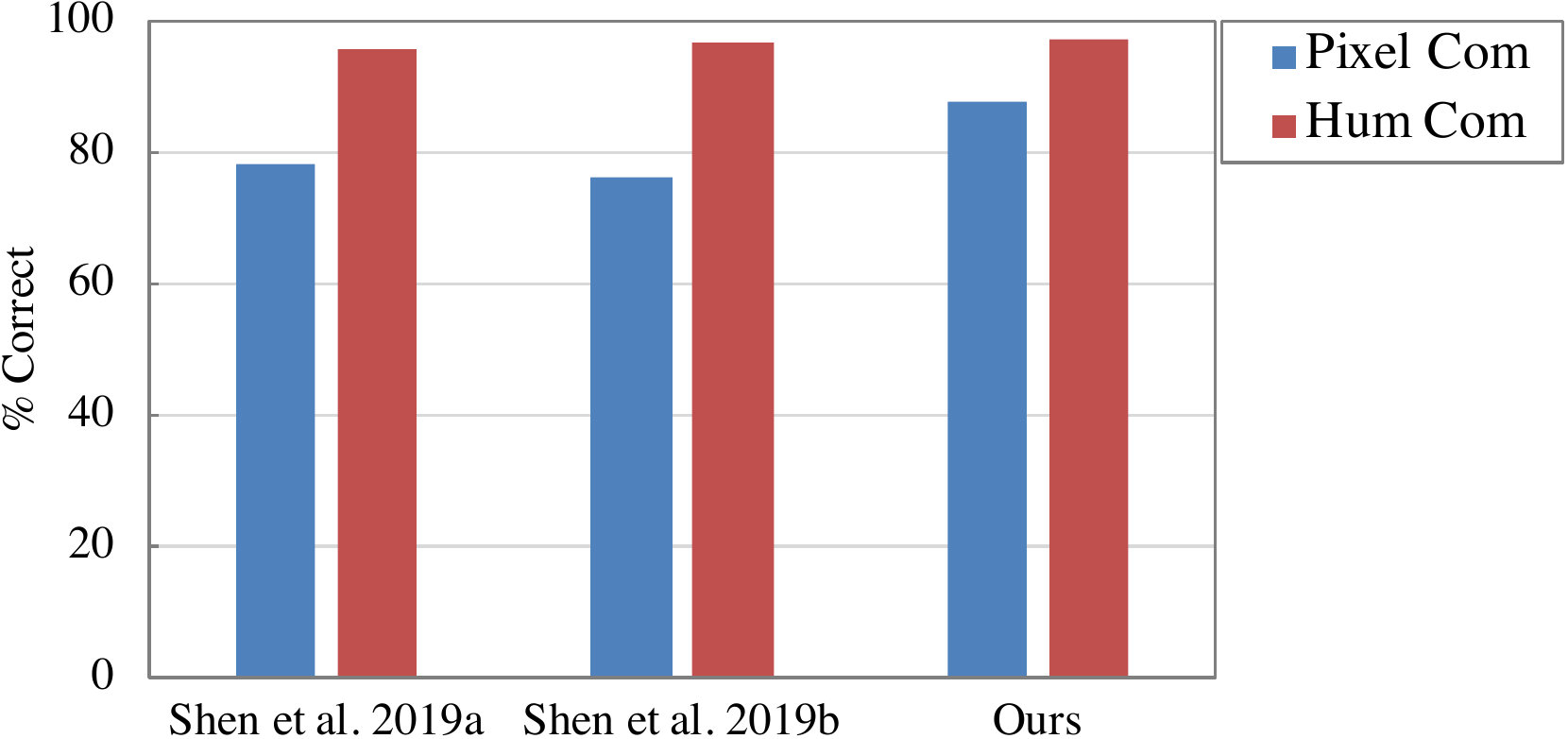}
	\end{center}
	\caption{Reconstruction quality of seen natural images by using different algorithms.}
	\label{fig:4}
\end{figure}

We also evaluate our method on complex stimuli (real color photos). Note that only very limited models can work on Dataset4. We compare our {\scshape{D-Vae/Gan}} with two recently proposed methods:~\cite{b1} and ~\cite{b2}. Since the evaluation is too complex for the natural images, following~\cite{b1,b2}, we conduct the comparisons by using Pix-Com and Hum-Com. As can be seen in Figure~\ref{fig:4}, our {\scshape{D-Vae/Gan}} achieves $87.8\%$ according to Pix-Com and $97.3\%$ according to Hum-Com, which significantly outperforms ~\cite{b1} by $9.7\%$ in Pix-Com and $1.6\%$ in Hum-Com, and surpasses~\cite{b2} by $11.7\%$ in Pix-Com and $0.3\%$ in Hum-Com. Some exemplars are visualized in Figure~\ref{fig7}.

\subsection{Ablation Studies.}

In order to validate the superiority of our {\scshape{D-Vae/Gan}} framework, we performed a series of ablation experiments in this section. 

\noindent {\bfseries \small Effectiveness of our {\scshape{D-Vae/Gan}}.} 
To quantify the contribution of our {\scshape{D-Vae/Gan}}, we derive a baseline, \emph{i.e.,} a standard DCGAN. On Dataset 1 and Dataset 3, DCGAN is very unstable in convergence, and thus unable to generate meaningful results for comparisons. On Dataset 2, as can be seen in Table~\ref{tab:2}, the reconstruction accuracy is quite dissatisfactory. We find our {\scshape{D-Vae/Gan}} brings significant performance improvements.

\noindent {\bfseries \small {Dual {\scshape{Vae}}-Based Encoder}.} 
Dual VAE-Based encoder is one of the major designs in this work, which forces the cognitive encoder to mimic the visual encoder. To evaluate this design, we offer a standard {\scshape{Vae/Gan}} model \emph{w/o} the paired visual encoder. It can be observed a great performance degradation after excluding {dual {\scshape{Vae}}-Based encoding model} (see Table~\ref{tab:2}).

\noindent {\bfseries \small {Standard Encoder} vs. {\scshape{Vae}}-Based Encoder.}  
To further investigate the superiority of using {\scshape{Vae}} for feature learning in our framework, we replace {\scshape{Vae}} in our model with standard Convolutional Neural Networks (CNNs). From Table~\ref{tab:2}, we also observe a significant performance drop. 

\begin{figure}[!t]
	\begin{center}
		\includegraphics[width=0.97\linewidth]{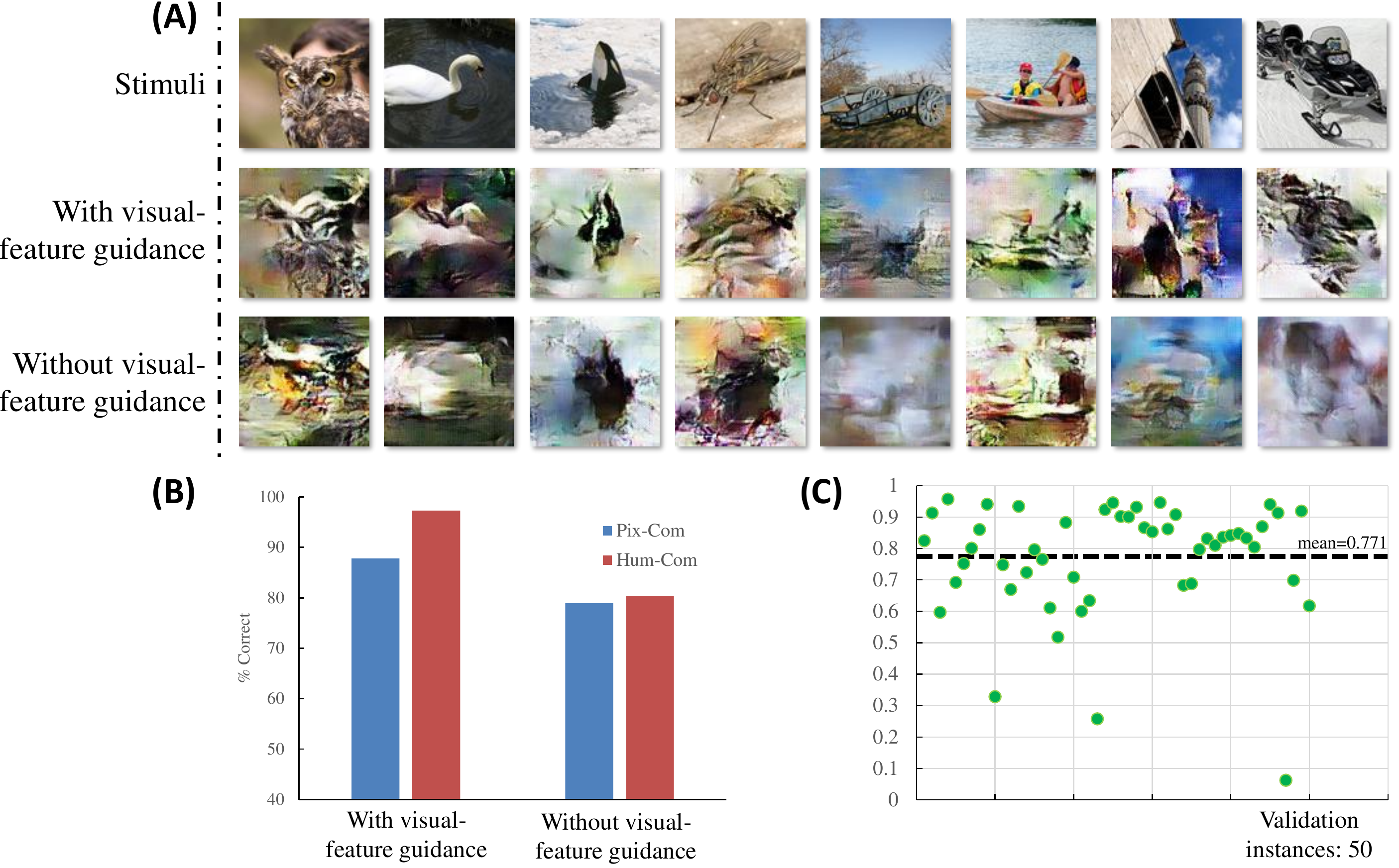}
	\end{center}
	\caption{Effect of visual-feature guidance. (A) Reconstructed visual stimuli \emph{w} $\&$ \emph{w/o} our visual-guidance strategy. (B) Reconstruction performance evaluated by Pix-Com and Hum-Com. (C) PCC between the learned cognitive features and the corresponding visual features of different instances.}
	\label{fig10}
\end{figure}

\noindent {\bfseries \small {Impact of Visual-Feature Guidance}.}
To investigate the effect of visual-feature guidance operated in this work, we also provide another solution that can also capture visual knowledge for comparisons. Specifically, we first trained a standard {\scshape{Gan}} based on a lot of natural images (no overlap with the test image set) such that the model learned visual information. Then with the generator fixed, this {\scshape{Gan}} model was combined with an encoder (same structure as our cognitive encoder) and trained to reconstruct visual stimulus from fMRI data \emph{w/o} knowledge distillation. As shown in Figure \ref{fig10}, our model achieves better performance than the model \emph{w/o} visual-feature guidance (\emph{i.e.,} $87.8\%$ vs. $78.9\%$ according to Pix-Com and $97.3\%$ vs. $80.3\%$ according to Hum-Com). 

\begin{figure}[!t]
	\begin{center}
		\includegraphics[width=0.98\linewidth]{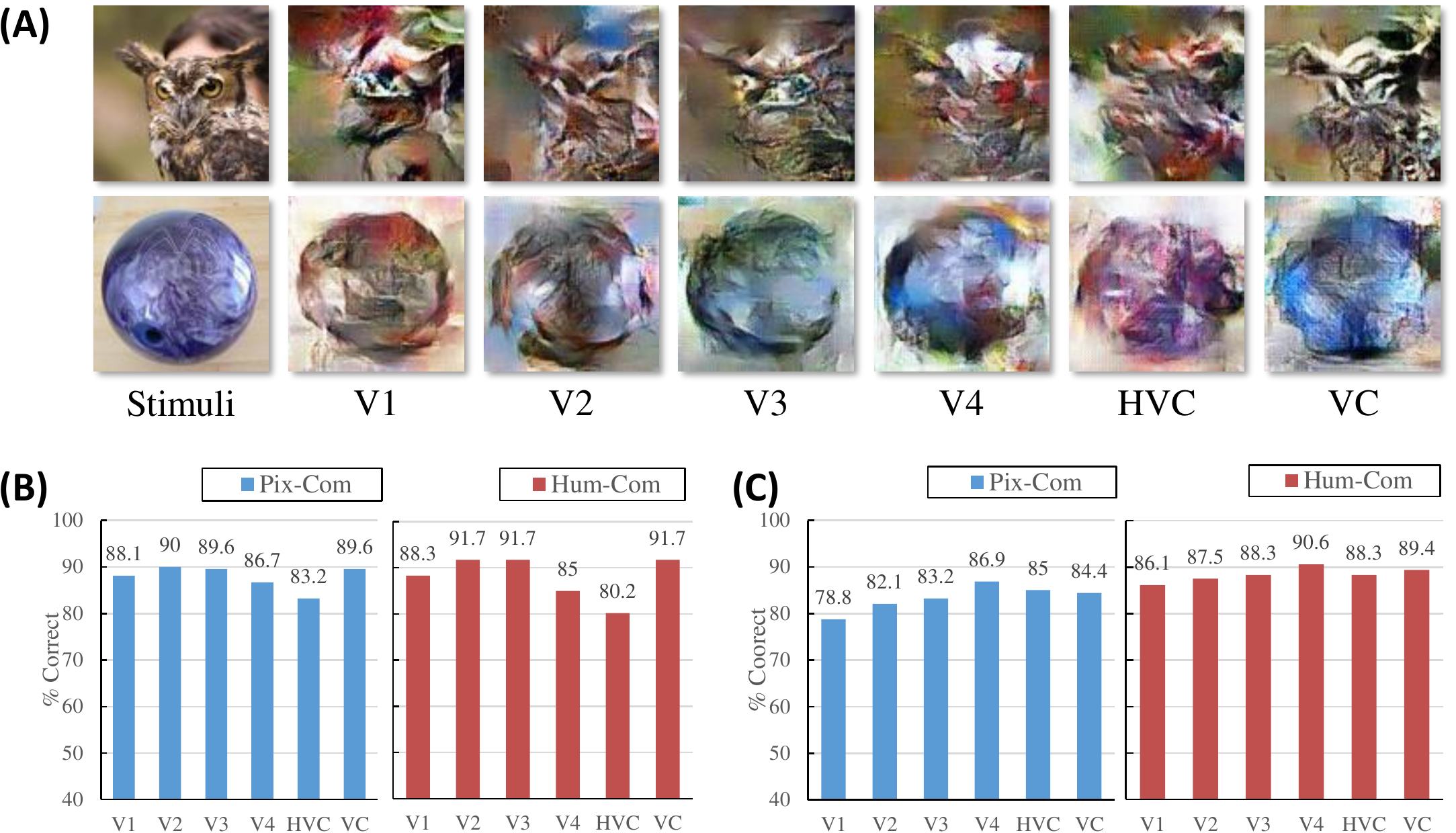}
	\end{center}
	\caption{{Reconstruction performance in different ROIs. (A) Reconstructions of natural images from multiple visual areas. (B) Reconstruction quality of structure for different visual areas. (C) Reconstruction quality of color for different visual areas.}}
	\label{fig11}
\end{figure}

\subsection{Reconstruction Performance in Different ROIs.}

To assess how the structures and colors of the stimulus images are reconstructed, we investigate the reconstruction quality for \emph{structure} and \emph{color} of different visual areas on Dataset 4 including V1 -- V4 and Higher Visual Cortex (HVC; covering regions around LOC, FFA, and PPA). 

The Pix-Com and Hum-Com evaluations both suggest that structures are reconstructed better from early visual areas (Figure \ref{fig11} (B)), whereas colors are reconstructed better from the midlevel visual area V4 (Figure \ref{fig11} (C)). It reveals that most structure-relevant information is preserved in the low-level visual cortices (V1 -- V3), while the midlevel (V4) and high-level cortices (LOC, FFA, and PPA) are responsible for capturing more color-relevant information. Above observations are entirely consistent with the coupling hierarchical relationship between perceived information and human visual areas \cite{al2012top}. Moreover, by applying the fMRI signals recorded from the entire visual cortex (VC), the structural and color characteristics are both effectively recovered so that these contrasting patterns further support the success of structure and color reconstructions. Besides, it indicates that our method can be used as a tool to characterize the information content encoded in the activity patterns of individual brain areas for visualization.

\section{Conclusions and Future Work}

In this paper, we achieve an encouraging improvement for the brain-driven visual stimulus reconstruction. We propose a novel framework {\scshape{D-Vae/Gan}} and a new three-stage training method to better target visual stimulus reconstruction task. Extensive experiments confirm the effectiveness of the proposed method. We also conduct experiments to analyse how visual information is captured by the ROIs in brain, which is meaningful for the high-level AI research including neural architecture design and brain emulation.

{\small
\bibliographystyle{ieee}
\bibliography{egbib}
}

\end{document}